\documentclass[letterpaper, 10pt, conference]{ieeeconf}

\IEEEoverridecommandlockouts                              
\usepackage{cite}
\usepackage{amsmath,amssymb,amsfonts}
\usepackage{algorithmic}
\usepackage[ruled,vlined,linesnumbered]{algorithm2e}
\usepackage{graphicx}
\usepackage{textcomp}
\usepackage{xcolor}
\usepackage{todonotes}
\usepackage{nicefrac}
\usepackage{comment}
\usepackage{multirow}
\usepackage{makecell}
\usepackage{bm,times}
\usepackage{marginnote}
\usepackage{siunitx}
\usepackage{url}

\title{\LARGE \bf
Solving Robot Assembly Tasks by Combining Interactive Teaching and Self-Exploration
}

\author{Mariano Ramirez Montero$^\dagger$, Giovanni Franzese$^\dagger$, Jeroen Zwanepol$^\dagger$, and Jens Kober
\thanks{$\dagger$ Equal contribution}
\thanks{The robot platform was kindly provided by Franka-Emika for the ERF competition and the writing of this article. 
This research is partially funded by European Research Council Starting
Grant TERI “Teaching Robots Interactively”, project reference 804907.}%
\thanks{The authors are with Cognitive Robotics, Delft University of Technology, The Netherlands (e-mail:\{g.franzese, j.kober\}@tudelft.nl, \{m.ramirezmontero-1, j.m.zwanepol\}@student.tudelft.nl).}%
}

\begin{document}
\maketitle
\begin{abstract}
Many high precision (dis)assembly tasks are still being performed by humans, whereas this is an ideal opportunity for automation. This paper provides a framework which enables a non-expert human operator to teach a robotic arm to do complex precision tasks. The framework uses a variable Cartesian impedance controller to execute trajectories learned from kinesthetic human demonstrations. Feedback can be given to interactively reshape or speed up the original demonstration. Board localization is done through a visual estimation of the task board position  and refined through haptic feedback. Our framework is tested on the Robothon benchmark disassembly challenge, where the robot has to perform complex precision tasks, such as a key insertion. The results show high success rates for each of the manipulation subtasks, including cases when the box is in novel poses. An ablation study is also performed to evaluate the components of the framework. 
\end{abstract}



\section{Introduction}


Robots can be stronger, faster and more agile than humans. However, humans are still much better at performing high precision assembly tasks. In fact, they are gifted with incredibly accurate sensors, i.e., visual and tactile, and extremely dexterous hands \cite{okamura2000overview}. Moreover, their brain is able to better handle many different and uncertain situations compared to robots \cite{liu2019human}. Activities like inserting a key in a keyhole or opening a door may be trivial for a toddler, but challenging for a robot. However, in a fast growing society where high precision manufacturing is still performed by underpaid workers, an automation revolution is necessary. Although a static object with fixed position and orientation is easy to automate, this becomes a lot harder when the object can be in a random position or orientation. This need for high precision manipulation can be seen in more parts of society, such as food production and logistics. 

Moreover, with more and more companies focusing effort on limiting waste and improving their sustainability, the demand for high precision disassembling tasks of old electronics will likely grow. It is currently not economically convenient for companies to focus on recycling their electronic waste manually, given the necessary allocation and training of more workers. 
However, recent advances on perception systems, robot control, and machine learning algorithms are presenting a possible fast robotics solution to this problem. Imagine if a single human could train many robots to perform complex assembly tasks and correct them when mistaken. 
Teaching robots through demonstrations and correction is the topic of interactive imitation learning (IIL). It has shown promising results on real robot applications, proving an alternative to pure reinforcement learning (RL), where the robot is learning from trial and error and may require long and dangerous interaction with the environment before succeeding, in particular when dealing with manipulation tasks \cite{rajeswaran2017learning} \cite{voigt2021multi}. 
\begin{figure}[t!]
    \centering
    \includegraphics[width=0.80\linewidth]{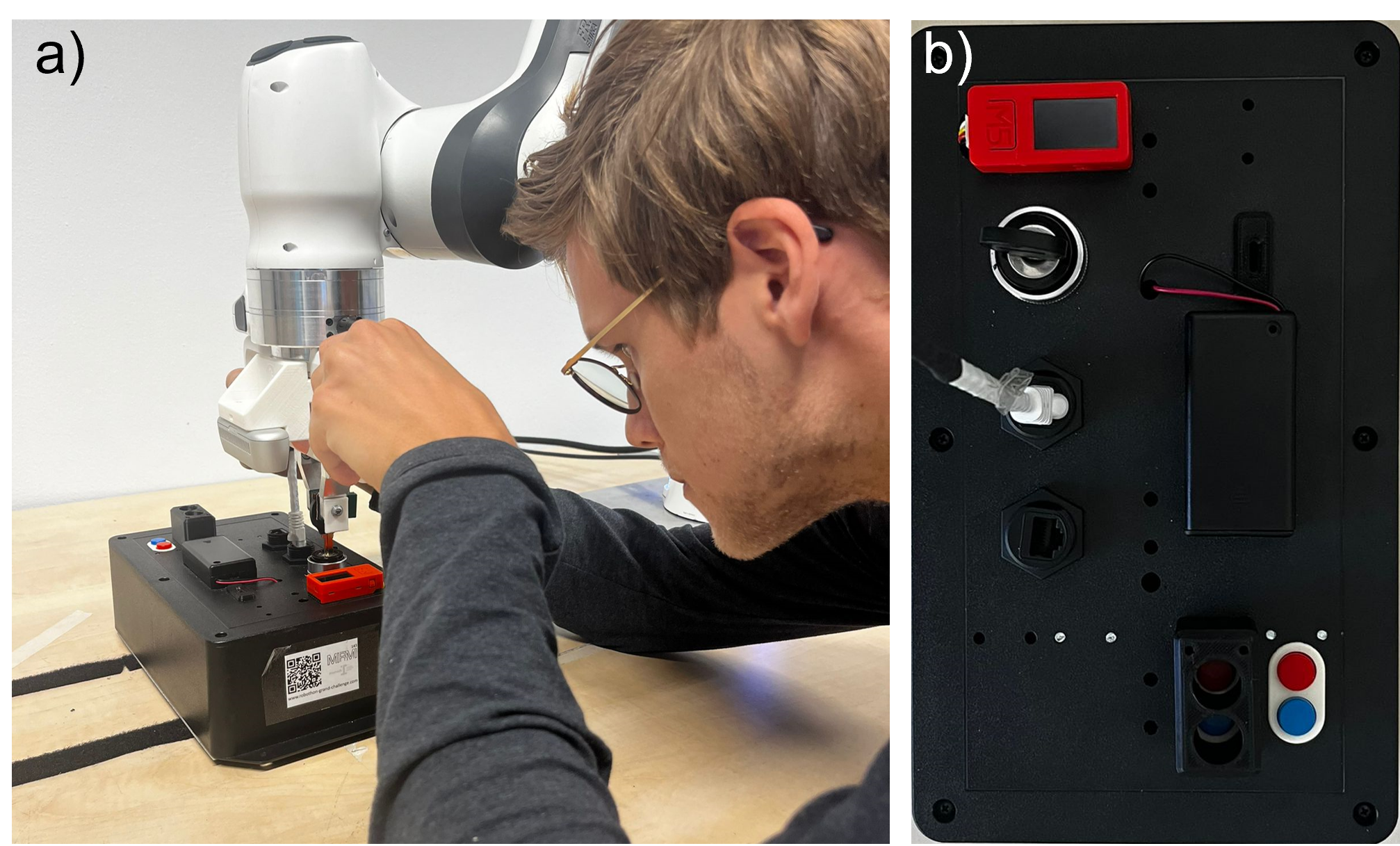}  
    \caption{a) Recording demonstration with kinesthetic teaching, b) Robothon task board \cite{robothon}}
    \label{fig:cover}
\end{figure}
This article contributes to the research of this topic with an interactive framework to teach robots impedance policies from human demonstration and corrections and it is able to perform high precision extraction and insertion tasks, i.e., removing batteries from their case and placing them in the appropriate recycling deposit. The proposed framework is able to locate the electronic board using visual inputs while improving the localization accuracy using haptic feedback. The robot control does not rely on any path planners or hard coded poses, instead it purely learns from human kinesthetic demonstration and teleoperated feedback. Moreover, to increase the reliability of the insertion task, an auxiliary searching strategy increases the success rate of the bare execution of the recorded human policy, handling the uncertainty on the electronic board location. 
The experiments are conducted with a 7 DoF Franka-Emika Panda and solve the same tasks proposed for the Robothon grand challenge \cite{robothon}. An initial stage of this strategy was presented at the European Robotics Forum's hackathon (Rotterdam, 2022) and won the competition out of the 4 teams participating. This article shares more details on the implementation of the strategy and performs a systematic evaluation of the framework including an ablation study and an analysis of particularly complex edge cases. The code of the complete framework can be found at \url{https://github.com/maxspahn/manipulation_tasks_panda} .

\section{Related Works}



Imitation learning methods such as DAgger\cite{dagger} have shown that robots can learn efficiently from expert demonstrations. However, providing error-free demonstrations is challenging and requires multiple trials and skilled human demonstrators. To this end, methods such as in  \cite{kelly2019hg}, \cite{franzese2021ilosa} propose to learn the robot tasks by human demonstrations and to refine the motion with corrective feedback. Giving demonstrations in the form of corrective feedback is more user-friendly for non-expert users and can even yield better robot performance than repeating full demonstrations \cite{perez2018interactive}.
These methods can be referred to as interactive imitation learning (IIL) techniques.
In the context of assembly tasks, \cite{franzese2021ilosa} tested IIL on a plugin task, i.e., two pegs in two holes, with non-expert users that first gave a kinesthetic demonstration and then provided incremental correction with a 3D-space mouse on the desired policy of the robot. The users managed to teach the task in only a few iterations. However, the socket was never moved from its original position. In this article, we developed the ideas presented in \cite{franzese2021ilosa}, but we tackle the challenge of even more complex insertion tasks and with the added uncertainty of (unreliable) localization. 

To handle the uncertain localization of the hole when performing insertion tasks, \cite{jha2022imitation} proposes to learn a model from demonstration with Gaussian Process (GPs) that maps the recorded external force with a desired displacement of the end effector. However, the tasks require many demonstrations and it is validated on a peg-in-the-hole insertion with high tolerances. Additionally, training a prediction model on the desired displacement to find the insertion point only based on external force can become ambiguous: colliding with the edges will generate a vertical force without any information on the right location of the hole. Alternatively, our framework also uses the external force, but instead to trigger an exploration policy that will increase the likelihood of finding the hole. This strategy is general and can be reused for different insertion tasks, i.e., key, battery, Ethernet cable, without the need of training an ad-hoc model per task. 

A combination of impedance control with a residual reinforcement learning policy was proposed in \cite{kulkarni2022learning},  
where a recurrent neural network (RNN) is trained with residual RL to predict the desired displacement given the external force input. The resulting policy is a vibrating exploratory policy.
The framework showed promising results but was only tested on a simple toy example, such as the assembly of the Hanoi tower, and it seems to learn how to approach only from one side of the hole. Alternatively, \cite{luo2019reinforcement} trained a NN architecture on a similar task of the peg-in-the-hole but concluded that it is not convenient to concatenate position and force information as input for the policy of the goal position calculation: when the object is displaced a few centimeters, the robot attempts dangerous behaviours. Otherwise, \cite{beltran2020variable} proposes a similar architecture but is evaluated only with a maximum uncertainty of the hole position of 1 mm, which in our framework would likely not require any activation of the self-exploration. Moreover, this is not compatible with the larger uncertainties given by the vision systems.

In the context of variable impedance learning, a recent work \cite{oikawa2021reinforcement} proposed to use RL for the automatic tuning of the non-diagonal stiffness matrix, but used an originally hand-coded policy for the sliding  peg-in-the-hole task. In our experience, we observed that a diagonal stiffness behaves well enough even in complex assembly tasks, and that the critical part is the learning of the attractor policy over the stiffness policy. 




%
To address the lack of interpretability of NNs, a systematic multi-level approach for the search of the hole is proposed in this framework
\cite{johannsmeier2019framework}, and recently in \cite{voigt2021multi}. 
The primitives for the insertions are hand-coded, i.e., the robot is always sliding the peg from one direction and the algorithm is using black-box optimization for tuning the hyperparameters at each iteration until performing a successful insertion. 
Similarly to this multi-level approach, our proposed framework retains the highest possible interpretability on what is happening during execution, while preserving flexibility and generalizability of the framework. 


Many different publications are testing the performance on lab-designed tasks or simple toy examples where no real precision is actually required, and usually the results are not reproducible. For this reason, we decided to benchmark our strategy on an official manipulation competition. 
Similarly, a recent work \cite{wittmann2022robotic}, using the same robot and benchmark, describe their successful use of a task scheduling strategy, path planning, and robot control. However, all the necessary via-points are hard coded and no interactive corrections can be used to locally reshape the motion or increase the robot speed. Our framework aims to make the teaching of  \emph{any} assembly tasks easier for non-expert users, not only to solve this specific competition. 

\section{Interactive Teaching Framework }
\begin{figure*}[t!]
    \centering
    \includegraphics[width=\linewidth]{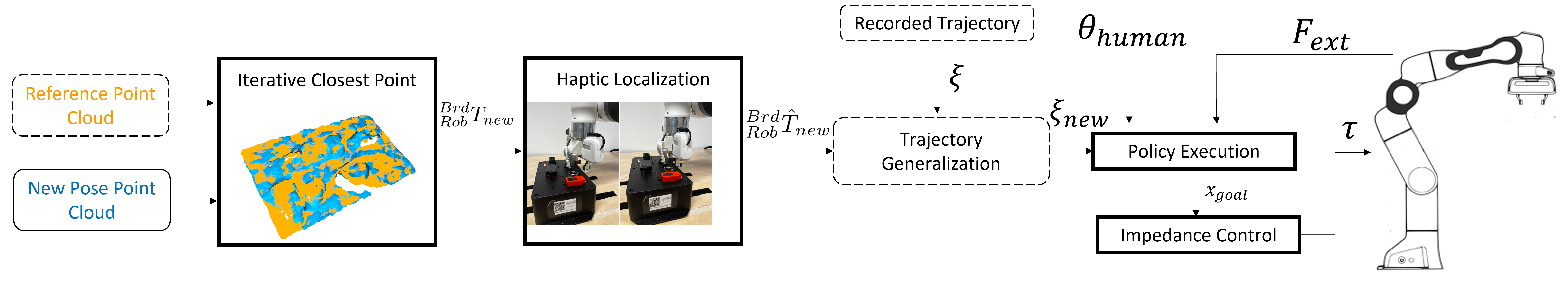}
    \caption{Board localization and task execution scheme}
    \label{fig:pipeline}
\end{figure*}

The goal of the proposed framework is providing the simplest possible Cartesian control, avoiding reliance on computationally heavy modules like inverse kinematics or path planning algorithms as proposed in many other works performing complex assembly tasks. 
The proposed framework can be divided in 3 main sub-groups: the Localization module (see Sec.~\ref{sec:localization}), the trajectory learning and execution module (see Sec.~\ref{sec:learning})  and the robot control module (see Sec.~\ref{sec:control}). Fig.~\ref{fig:pipeline} depicts the different components of the framework. 
The following section explains the modules starting from the variable impedance controller which allows to switch from execution and human-in-control. Then the trajectory learning from human demonstrations, interactive corrections and self-exploration which allows for fast teaching of the robot while being reliable with the uncertainty of insertion tasks is explained. We finish with the localization strategy that allows for the generalization of the learned policy to different locations of the object to assemble. 

\subsection{Variable Cartesian Impedance Control}\label{sec:control}
The dynamic equation of the robot is defined according to
\begin{equation}
    \bm{\mathcal{M}}(\bm{q}) \ddot{\bm{q}}+\bm{\mathcal{C}}(\bm{q},\dot{\bm{q}})+\bm{\mathcal{G}}(\bm{q})= \bm{\tau}_{task} +\bm{\tau}_{NS} + \bm{\tau}_{lim}+ \bm{\tau}_{ext} 
\end{equation}
where, in order from left to right, there are the mass, the Coriolis and the gravitational term that depend on the joint configuration $\bm{q}$ and, on the right, the torque for the Cartesian (or task) control, the nullspace control torque, the repulsion from joint limit torque and finally the externally applied torque. 
The task space torque is defined as
\begin{equation}
    \bm{\tau}_{task}=\bm{{J}}^\top(\bm{\mathcal{K}}(\bm{x}_{goal}-\bm{x})-\bm{\mathcal{D}\dot{\bm{x}}})+\bm{\mathcal{C}}(\bm{q},\dot{\bm{q}})+\bm{\mathcal{G}}(\bm{q})
\end{equation}
where $\bm{{J}}$ is the geometric Jacobian, and the stiffness $\bm{\mathcal{K}}$ and the damping $\bm{\mathcal{D}}$ give the compliant behaviour with a critically damped response \cite{albu2003cartesian}. $\bm{x}_{goal}$ is the goal Cartesian pose of the end effector and $\bm{x}$ is the current Cartesian pose of the end effector. The compensation of the gravitational ($\bm{\mathcal{G}}(\bm{q})$) and the Coriolis terms ($\bm{\mathcal{C}}(\bm{q},\dot{\bm{q}})$) allows to have the robot not fall under its own weight when the human is performing kinesthetic teaching.

On the other hand, when working with a redundant manipulator,  null-space control can be formulated as a projection of the joint impedance in the kernel of the end effector's Jacobian \cite{albu2003cartesian}, according to 
\begin{equation}
    \bm{\tau}_{NS}=(\bm{I}-\bm{J}^\top \bm{J}^{\top +}) (\bm{\mathcal{K}}_{NS}(\bm{q}_{NS}-\bm{q})-\bm{\mathcal{D}}_{NS}\dot{\bm{q}}).
    \label{eq:nullspace}\end{equation}
where $\bm{\mathcal{K}}_{NS}$ is the null space stiffness matrix, $\bm{\mathcal{D}}_{NS}$ is the null space damping matrix, and $\bm{q}_{NS}$ is the desired configuration. 
This creates a lower priority on the desired joint configuration that will only generate joint transitions that have a minimal effect on the imposed end effector dynamics. 
Additionally, the same rule can be used to minimize the risk of hitting a joint limit. We define $\bm{q}_{safe}$ as the joint configuration with a safety distance from the joint limit $\bm{q}_{lim}$. In case the robot joint configuration $q$ surpasses the safety limit $\bm{q}_{safe}$, a null space controller is activated with $\bm{q}_{NS}=\bm{q}_{safe}$ only for the joints which are beyond their safety limit. 
Moreover, the stiffness and the damping for the joint limit rejection are non-zero only when the robot goes beyond the safety limit. It is important that this limit is set before the real-joint limit, otherwise the robot's safety control will lock the joints. 

The projection of joint limit torque in the null-space of the Cartesian control generates transitions in the robot configurations that move every joint away from its limit while not interfering with the desired pose control that is actuated in the Cartesian space. This simple solution allows the robot to physically converge to kinematically feasible solutions (when existing) without the need of any inverse kinematics controllers or planning which can be unreliable and converge to local minima. 
In the context of assembly tasks, when performing variable impedance control to acquire human demonstrations or decrease the insertion force, the rate of change of stiffness cannot be too drastic. As shown in \cite{kronander2016stability}, a too fast increase of the stiffness can generate dangerous robot instability. For this reason, the stiffness modulation is done with a proportional controller with respect to the requested target one, i.e. $\dot{K}=\alpha ({K}_{target}- K)$. 

\begin{algorithm}[t!]
\SetAlgoLined
\DontPrintSemicolon
\textbf{a) Kinesthetic Teaching} \\
t = 0  
\While{Trajectory Recording}{ 
	{Receive($\bm{x} ,t$)} \\
    $\mathcal{D} \leftarrow \mathcal{D} \cup (\bm{x}, t);$ \\
    $t = t +\Delta t$
 } 
$\text{Train}(\pi_{robot}) \ \ \text{with} \ \ \mathcal{D}$\\
\textbf{b) Interactive Learning with Self-Exploration} \\
t = 0 
\While{Control Active}{
	{Receive($\bm{x}$}) \\
    
\If{Received Human Feedback}{
    $\pi_{robot} \leftarrow \pi_{robot} +  \pi_{human}$ \\
    }
\If{Exploration Triggered}{
    $\pi_{robot} \leftarrow \pi_{robot} +  \pi_{explore}$ \\
    }
$\bm{x}_{goal} = \pi_{robot}$ \\
Send ($\bm{x}_{goal}$)\\
$t = t + \Delta t $ \\
}
\caption{Interactive Robot Learning}
\label{algo::interactive_teaching}
\end{algorithm}
\subsection{Trajectory Learning}\label{sec:learning}
\paragraph{Kinesthetic Demonstration}The policy of the robot arm is learned through kinesthetic demonstration, which can be seen in lines 1-6 in Algorithm~\ref{algo::interactive_teaching}. The demonstration of each task is given by manually moving the end effector of the robot to the desired positions, while closing and opening the gripper at the desired times. In order for the robot to be compliant enough to give such demonstrations, the stiffness of the end effector is set to zero in all Cartesian directions but the nullspace joint limit is kept active to assist the demonstrator in handling the complex kinematic chain of the robot. The Cartesian pose of the end effector is \emph{aggregated} in the database, denoted as $\mathcal{D}$ in the algorithm, for each time step and then used to train the policy for each of the subtasks.

The fitting of the trajectory, i.e., $\bm{x}=f(t)$ can be performed independently in every Cartesian direction using the desired regressor, i.e., linear or cubic spline, Gaussian Processes (GPs) \cite{huang2019kernelized}, or Dynamic Movement Primitives (DMPs) \cite{saveriano2021dynamic}. However, even if the interface with different approximation functions is possible, during testing and development a linear spine is used to prove that the effectiveness of the framework does not need to rely on the use of advanced function approximators. 

\paragraph{Trajectory Segmentation}When performing a complex assembly task like the one in Fig.~\ref{fig:cover}, composed of multiple simpler tasks, the user can stop after the execution of each of them and save it as a movement primitive for that specific task. Additionally, the demonstrator can also label the task as an insertion task or not. In case of an insertion task, during the execution, if the robot is \emph{measuring} a high contact force in the direction of insertion, the self-exploration policy is activated. This is desired because when a high contact force is registered, this is most likely because the robot has not found the object's slot.

Similar works like \cite{manschitz2020learning}, automatically segment the motion and the control modality, i.e., force/position. However, since we do not employ any external force sensor and have to deal with imperfect human demonstrations, we prefer to teach every segment separately, to avoid over/under segmentation and to explicitly label whether each segment is an insertion task. This would be complex to infer only from recorded data and would most likely be prone to misclassification. 
At the start of each subtask execution, the robot will move from the end pose of the previous primitive to the start of the current primitive with a constant velocity before reproducing the learned behaviour. 

\paragraph{Interactive Trajectory Shaping}However, when learning a policy on top of an impedance control, the result of the execution cannot always match with the intended human behaviour. For example, when pressing a button, or sliding out the battery cover, a high contact force is required. An elegant solution is to interactively shape the attractor trajectory of the end effector using teleoperated human feedback, as also proposed in \cite{franzese2021ilosa}. Given the feedback, the attractor is moved under the contact surface and, thanks to the set robot impedance, a higher contact force is obtained which can be used for pushing, pulling or generating enough contact force for opening a battery case.  
In the proposed framework, the feedback to the reference trajectory is applied by a human using directional feedback like in \cite{perez2018interactive}, using a keyboard. This means that the teacher will give incremental \emph{corrections} in the desired direction and the magnitude is set as a hyperparameter. 
However, the feedback should not have an effect only on the current position of the trajectory but needs to be spread smoothly on all the points of the trajectory, to avoid overfitting and maintain smoothness. For this reason, the update rule makes use of an infinite-smooth function, i.e., a squared exponential, that computes the feedback magnitude of each element of the database based on their Euclidean distance from the current robot position, according to 
\begin{equation} 
    \bm{\hat{\xi}} = \bm{\xi} +\bm{\theta} \exp
    \left(-\frac{\Vert{\bm{\xi} -\bm{x}}\Vert^2}{l^2}\right)
    \label{eq:squared_exp}
\end{equation}
where $\bm{\xi}$ is the original reference trajectory, $\bm{\hat{\xi}}$ is the new reference trajectory after the feedback, $\bm{x}$ is the current position of the end effector, $\bm{\theta}$ is the vector with the feedback in Cartesian axis and $l$ is a length scale hyperparameter. For larger length scales, the feedback will affect more points in the neighborhood of the current point of the trajectory.

Another caveat of using kinesthetic teaching for this high precision task is that demonstrations from the human operator might be performed slower than the desired robot execution. Increasing the speed of these slow demonstrations can be very beneficial, especially in the case of some task being repeated very frequently. However, increasing the speed uniformly may result in undesired high interaction forces or impacts with the environment, for example when bumping  on a bottle during picking \cite{meszaros2022learning}. For this reason, it is once again desirable to have the human operator give feedback to the robot for speeding up the demonstration locally, since the human operator can decide what are safe/desirable speeds at each particular time of the motion. When the user is requesting to go faster, the rate of the motion is increased $n$-times in the following $m$ seconds of the motion. From an algorithmic point of view, given the discrete nature of the trajectory, we uniformly remove $ \lfloor{ m / \Delta t(1- 1/n) \rfloor} $ points from the trajectory in the next $m$ seconds of it. Since we used incremental feedback, we manually set $m$ to $0.2$ seconds and $n$ to 2. 
This means that, given the controlling $\Delta t$ of $0.01$ seconds, when the feedback button is pressed, 10 points are removed uniformly from the next 20 points in the trajectory. This uniform \emph{dis-aggregation} of data allows making the motion faster without the generation of discontinuities in the trajectory that could make the motion shaky or even unstable. 

\paragraph{Exploration Policy}Nevertheless, when learning complex manipulation operations like the insertion of a key, trajectory execution and correction is not enough anymore to guarantee a successful repeatability of the task. Therefore, if the movement primitive is labelled as an insertion task and if the vertical force goes beyond an activation threshold, an exploration policy is activated until the contact force decreases below a deactivation threshold, since this likely indicates the insertion was performed successfully. 
Any exploratory policy could be used, i.e., uniform, Gaussian, Brownian noise or concentric circles, Lissajous and spiral paths \cite{jiang2022review}.
We conducted all our experiments using the spiral path, line 14-16 of Alg. \ref{algo::interactive_teaching} .
It is noteworthy to remark that increasing the Cartesian stiffness to reduce the tracking error is not an effective solution since the object localization has irreducible uncertainties that will make the insertion fail and consequently generate enormous forces on the components \cite{peternel2018robotic}. However, reducing this uncertainty to the smallest possible value is necessary considering that the self-exploration strategy is only acting locally and will fail if the localization error is larger than the exploration radius, or another \emph{hole} is found. The following chapter will focus on localizing the object using both visual and haptic inputs. Furthermore, the method for transforming the learned policy with respect to the new object location will be explained. 

\subsection{Object Localization}\label{sec:localization}

As stated in the previous section, for such high accuracy insertion tasks it is necessary to reduce the localization uncertainties as much as possible. Since attempted visual localization approaches did not prove to be accurate or robust enough, a combined approach was taken where visual localization provides an initial guess for the haptic localization.

Differently from \cite{bjorkman2013enhancing} or \cite{dragiev2011gaussian}, here the haptic and visual feedback are used to reconstruct the object shape. We have the hypothesis of already knowing the object shape and use the visual feedback to determine an initial guess of the change of location in the scene to subsequently refine it with haptic interactions. While this is a significant assumption that also limits generalization of our object localization approach, it is worth noting that for this case it is only necessary to record an initial reference point cloud and haptic localization of the board where the kinesthetic demonstration are provided. 
This is much more data efficient than a (deep) learning approach which might need a lot of learning data \cite{wittmann2022robotic}. Moreover, while these deep learning approaches might be able to better generalize to harder situations (with occlusions, for example), the uncertainty in state-of-the-art methods is still likely too large for assembly task. A typical evaluation metric is `5cm 5deg' \cite{shotton20136dposedeep}, meaning a pose is considered correct if it is within 5cm in translation and 5 deg in rotation, while these tasks require accuracies under {1}cm in translation. 
\newcommand{\tr}[4]{ {}^{#2}_{#3} {#1}_{#4}  }
Concretely, the visual localization of the box uses this set of transformations
\begin{equation}
    \tr{T}{Brd}{Rob}{new} = \tr{T}{Cam}{Rob}{} \ \tr{T}{}{}{ICP} \ \tr{T}{Rob}{Cam}{} \ \tr{T}{Brd}{Rob}{start} 
    \label{eq:tranform}
\end{equation}
where, starting from right, the original transformation of the board in the location of the recording is translated in the camera frame, then transformed according to the colored Iterative Closest Point (ICP) \cite{zhou2018open3d}, and subsequently re-translated in robot frame. Given this new estimation of the board, two collision trajectories are sent on the sides of the board with zero rotational stiffness in the z-axis, letting the end effector align with the board's sides, see Fig.~\ref{fig:pipeline}. The two lines defined by positions and orientations of the robot after collision with the board are intersected to update the board localization in the robot frame to  $\tr{\hat{T}}{Brd}{Rob}{new}$ with higher accuracy. 
Finally, the original demonstration, recorded in the robot frame, is projected first in the coordinate of the board at the start and then projected back in robot frame using the new transformed frame, according to
\begin{equation}\label{eq:transform_traj}
    \bm{\xi}_{new} = \tr{\hat{T}}{Brd}{Rob}{new} \ \tr{T}{Rob}{Brd}{start} \ 
\bm{\xi}.
\end{equation}

\section{Real Robot Validation Experiments}

This sections summarizes and discusses the results obtained from testing the framework and its components using the Franka-Emika Panda robot on the Robothon task board. The tasks are: push two buttons; remove an Ethernet cable from a socket and plug it
into another; grasp a key, insert it into a keyhole and turn it; remove two batteries from a case and put them in a storage, see Fig.~\ref{fig:cover}.b).
In order to increase the grasping consistency of the robot, we 3D printed fingertips with a concave shape, to better grasp the battery, next to a flat shape, to grasp everything else. 
The board was located in front of the robot like in Fig.~\ref{fig:cover} and after saving the point-cloud model and the refined frame with haptic feedback, the user is asked to perform kinesthetic teaching and corrections. A video of the experiments can be found here \url{https://youtu.be/qYujdzYhjAo}. 

\subsection{Interactive Correction Evaluations}

\begin{figure}[t!]
    \centering
    \includegraphics[width=0.4\textwidth]{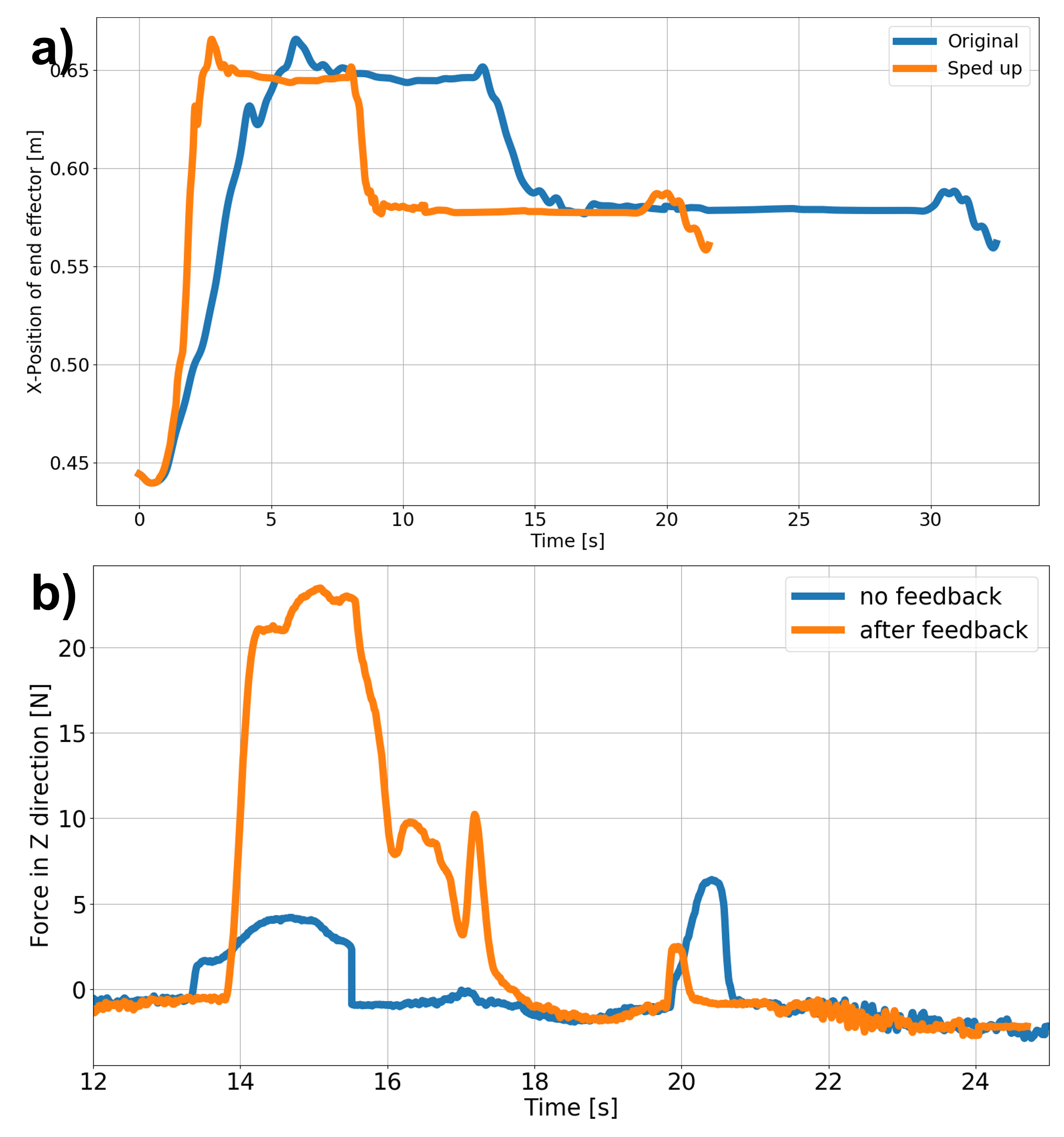}
    \caption{a) comparison of trajectory in x-axis for key insertion before and after speeding it up. b) change of contact force in the z-axis when opening the battery box before and after corrective feedback}
    \label{fig:corrections}
\end{figure}

To provide a visual intuition of how the corrections are necessary for increasing the performance of the demonstration, Fig.~\ref{fig:corrections} depicts in a) the effect to the speeding up feedback compared to the original demonstration when performing key insertion, while being able to successfully insert the key and in b) the change of contact force when opening the battery box before feedback (and failing to open it) and after feedback (with successful opening). Indeed, shifting of the attractors under the case surface (see Eq.~\eqref{eq:squared_exp}), increases the vertical contact force and the resulting friction of the gripper with the case, obtaining enough force to slide the case open. It is worth noting that without corrections, i.e., simply performing kinesthetic teaching, the user would not be able to demonstrate the desired high vertical force unless an external force sensor is mounted between the arm and the hand. 

\subsection{Task Generalization}
To test how well our approach generalizes, the proposed framework was tested on \emph{five} different poses of the task board, as well as in the reference pose where the demonstrations were performed. The success rates per task can be seen in Table~\ref{tab:success_rates}. \emph{Five} trials of the full task board were performed per pose. 
The consistent success of all the tasks in the different poses 1,2,3 show the robustness of the framework. In these poses only one failure occurred for the key insertion and for the ejection of battery 1, both in pose 3.

On the other hand, the success rates for each task calculated from the 5 trials for poses 4 and 5 are given separately, since these poses caused more failures on different tasks and thus are more interesting to look at separately. Pose 4 had a clockwise rotation of approximately $90^\circ$ with respect to the reference pose, which in turn made it difficult for the robot to reach the desired poses without reaching its joint limits. This meant that the key was picked at a different end effector orientation than in the demonstration, which made insertion much more difficult. Nonetheless, with self-exploration, insertion still succeeded 3/5 trials. Similarly, for the picking of the batteries, the joint limits of the robot were reached, in these cases, however, the robot fully stopped, failing to pick either of the batteries. Finally, the insertion of the Ethernet cable failed for the same reasons as the key. These were found to be configurations that did not have a feasible solution, independent of the controller used.  
Thus, many of these failures were the consequence of the limited reachability of the robot. Although the null-space rule of Eq.~\eqref{eq:nullspace} showed a successful adaptability of the kinematic chain for completely new poses, it could not overcome the limitation of asking for unreachable poses. Additionally, the joint limit rejection is still a myopic rule that is only adapting the kinematic chain locally to accommodate the requested pose. Future works should combine the rule with more advanced null-space path planning to possibly find feasible configurations, if such exist. 

\begin{table}[t!]
\centering
\caption{Success rates per task for the tested board poses}
\label{tab:success_rates}
\begin{tabular}{l|ccc|}
\cline{2-4}
                                                                                          & \multicolumn{3}{c|}{\textbf{Success rate}}                \\ \cline{2-4} 
 &
  \multicolumn{1}{c|}{\textbf{\begin{tabular}[c]{@{}c@{}}Reference, Pose 1, \\ Pose 2, Pose 3\end{tabular}}} &
  \multicolumn{1}{c|}{\textbf{Pose 4}} &
  \multicolumn{1}{l|}{\textbf{Pose 5}} \\ \hline
\multicolumn{1}{|l|}{\textbf{\begin{tabular}[c]{@{}l@{}}Blue Button\end{tabular}}}     & \multicolumn{1}{c|}{100\%} & \multicolumn{1}{c|}{100\%}   & 100\%\\ \hline
\multicolumn{1}{|l|}{\textbf{\begin{tabular}[c]{@{}l@{}}Key Pick\end{tabular}}}         & \multicolumn{1}{c|}{100\%}  & \multicolumn{1}{c|}{100\%}   & 100\%\\ \hline
\multicolumn{1}{|l|}{\textbf{\begin{tabular}[c]{@{}l@{}}Key Insert\end{tabular}}}       & \multicolumn{1}{c|}{95\%} & \multicolumn{1}{c|}{60\%} & 0\% \\ \hline
\multicolumn{1}{|l|}{\textbf{\begin{tabular}[c]{@{}l@{}}Battery  Case\end{tabular}}}    & \multicolumn{1}{c|}{100\%} & \multicolumn{1}{c|}{100\%}   &  100\% \\ \hline
\multicolumn{1}{|l|}{\textbf{\begin{tabular}[c]{@{}l@{}}Battery 1 Eject\end{tabular}}}  & \multicolumn{1}{c|}{95\%}  & \multicolumn{1}{c|}{0\%}   & 0\%\\ \hline
\multicolumn{1}{|l|}{\textbf{\begin{tabular}[c]{@{}l@{}}Battery 1 Pick\end{tabular}}}  & \multicolumn{1}{c|}{100\%} & \multicolumn{1}{c|}{0\%}   & 100\%\\ \hline
\multicolumn{1}{|l|}{\textbf{\begin{tabular}[c]{@{}l@{}}Battery 1 Place\end{tabular}}} & \multicolumn{1}{c|}{100\%} & \multicolumn{1}{c|}{0\%}   & 0\%\\ \hline
\multicolumn{1}{|l|}{\textbf{\begin{tabular}[c]{@{}l@{}}Battery 2 Eject\end{tabular}}} & \multicolumn{1}{c|}{100\%} & \multicolumn{1}{c|}{0\%}   & 100\%\\ \hline
\multicolumn{1}{|l|}{\textbf{\begin{tabular}[c]{@{}l@{}}Battery 2 Pick\end{tabular}}}  & \multicolumn{1}{c|}{100\%} & \multicolumn{1}{c|}{0\%}   & 0\%\\ \hline
\multicolumn{1}{|l|}{\textbf{\begin{tabular}[c]{@{}l@{}}Battery 2 Place\end{tabular}}} & \multicolumn{1}{c|}{100\%} & \multicolumn{1}{c|}{0\%}   & 0\%\\ \hline
\multicolumn{1}{|l|}{\textbf{\begin{tabular}[c]{@{}l@{}}Ethernet Pick\end{tabular}}}    & \multicolumn{1}{c|}{100\%} & \multicolumn{1}{c|}{100\%}   & 100\%\\ \hline
\multicolumn{1}{|l|}{\textbf{\begin{tabular}[c]{@{}l@{}}Ethernet Place\end{tabular}}}  & \multicolumn{1}{c|}{100\%} & \multicolumn{1}{c|}{0\%}   & 100\%\\ \hline
\multicolumn{1}{|l|}{\textbf{\begin{tabular}[c]{@{}l@{}}Red Button\end{tabular}}}      & \multicolumn{1}{c|}{100\%} & \multicolumn{1}{c|}{100\%}   & 100\%\\ \hline
\end{tabular}
\end{table}

\subsection{Ablation Study}


As an ablation study, we measured the success rate with and without haptic localization
for key and battery insertion tasks with a novel pose that is still within the robot's reachability. 
Thus for the case without haptic localization, only the transform from ICP is used to transform the trajectories, see Eq.~\eqref{eq:tranform}.
Table~\ref{tab:icp_ablation} shows the success rate and spiral search trigger rate over 5 trials for each of these cases. Firstly, it is worth noting \emph{spiraling is always triggered}, meaning the tasks could not succeed in any case without self-exploration. However, the low success rate of the battery insertion indicates that, as hypothesized, the visual localization by itself is too uncertain for such a task. As shown by the 100\% success rate in the haptic localization case, this second approach reduces the uncertainty enough to ensure success. 

\begin{table}[t!]
\centering
\caption{\centering Success rates for key and battery 2 insertion tasks with and without haptic localization}
\label{tab:icp_ablation}
\begin{tabular}{l|cc|cc|}
\cline{2-5}
 &
  \multicolumn{2}{c|}{\textbf{\begin{tabular}[c]{@{}c@{}}No haptic \\ localization\end{tabular}}} &
  \multicolumn{2}{c|}{\textbf{\begin{tabular}[c]{@{}c@{}}Using haptic \\ localization\end{tabular}}} \\ \cline{2-5} 
 &
  \multicolumn{1}{c|}{\textbf{\begin{tabular}[c]{@{}c@{}}Success \\ Rate\end{tabular}}} &
  \textbf{\begin{tabular}[c]{@{}c@{}}Spiraling \\ Rate\end{tabular}} &
  \multicolumn{1}{c|}{\textbf{\begin{tabular}[c]{@{}c@{}}Success \\ Rate\end{tabular}}} &
  \textbf{\begin{tabular}[c]{@{}c@{}}Spiraling \\ Rate\end{tabular}} \\ \hline
\multicolumn{1}{|l|}{\textbf{Key Insert}}      & \multicolumn{1}{c|}{100\%} & 100\% & \multicolumn{1}{c|}{100\%} & 100\% \\ \hline
\multicolumn{1}{|l|}{\textbf{Battery 2 Place}} & \multicolumn{1}{c|}{20\%}  & 100\% & \multicolumn{1}{c|}{100\%} & 100\% \\ \hline
\end{tabular}
\end{table}



\section{Conclusions}
This work presented an interactive framework based on impedance control and learning from demonstration for teaching high precision insertion tasks from non-expert users. The framework was tested on a manipulation benchmark and it was able to successfully solve all the subtasks, even when the electronic board was moved from the original position, with the aid of the proposed localization strategy which combines visual and haptic inputs to reduce the uncertainty as much as possible. Moreover, to deal with irreducible uncertainties, the use of a self-exploration searching strategy is necessary, and showed a drastic increase of the success rate. 
However, during testing we managed to find edge cases where the framework would fail to accomplish the tasks. We believe that, beyond the cases where it was not physically possible for the robot to reach the desired configuration, it could be beneficial to integrate our framework with long-horizon path planners and leverage the positive side of both of them.
Some limitations will affect the applicability of the framework to different scenarios, for example if the shape of the board is not rectangular, or it is not possible to apply high contact forces on the sides without moving, damaging or flipping the object. More sophisticated tactile sensors, such as the one employed in \cite{bjorkman2013enhancing}, could increase the applicability range of the localization with touch. Moreover, the searching strategy cannot be used when dealing with soft tissues, such as in surgery or when handling food, opening many different research directions as alternatives.  
Finally, the development of an active request for human feedback \cite{taylor2021active} could increase the reliability of the task execution on particularly hard edge cases, and help in learning a parametrization \cite{calinon2016tutorial} of the policy with respect to the current pose of the task board.
\pagebreak
\bibliographystyle{IEEEtran}
\bibliography{reference}  

\end{document}